\documentclass[conference]{IEEEtran}
\IEEEoverridecommandlockouts
\usepackage{amsmath,amssymb,amsfonts}
\usepackage{graphicx}
\usepackage{textcomp}
\usepackage{xcolor}

\makeatletter
\let\MYcaption\@makecaption
\makeatother

\usepackage[font=footnotesize]{subcaption}

\makeatletter
\let\@makecaption\MYcaption
\makeatother

\usepackage[margin=false, inline=true]{fixme}
\usepackage{algorithm}
\usepackage{algpseudocode}
\usepackage{multirow}
\usepackage[square,sort&compress,comma,numbers]{natbib}
\usepackage{amsthm}
\usepackage{url}

\renewcommand{\vec}[1]{\boldsymbol{\mathbf{#1}}}
\newcommand{\HLA}{\text{HLA}}

\usepackage{hyperref}

\def\BibTeX{{\rm B\kern-.05em{\sc i\kern-.025em b}\kern-.08em
    T\kern-.1667em\lower.7ex\hbox{E}\kern-.125emX}}

\begin{document}

\theoremstyle{definition}
\newtheorem{definition}{Definition}[section]

\title{A Latent Space Model for HLA Compatibility Networks in Kidney Transplantation\thanks{This research was partially conducted while the authors were at the University of Toledo.}}

\author{\IEEEauthorblockN{Zhipeng Huang}
\IEEEauthorblockA{Department of Computer and Data Sciences \\
Case Western Reserve University \\
Cleveland, OH 44106 USA \\
\url{zhipeng.huang@case.edu}}
\and
\IEEEauthorblockN{Kevin S. Xu}
\IEEEauthorblockA{Department of Computer and Data Sciences \\
Case Western Reserve University \\
Cleveland, OH 44106 USA \\
\url{ksx2@case.edu}}
}

\maketitle

\begin{abstract}
Kidney transplantation is the preferred treatment for people suffering from end-stage renal disease. 
Successful kidney transplants still fail over time, known as graft failure; however, the time to graft failure, or graft survival time, can vary significantly between different recipients.  
A significant biological factor affecting graft survival times is the compatibility between the human leukocyte antigens (HLAs) of the donor and recipient. 
We propose to model HLA compatibility using a network, where the nodes denote different HLAs of the donor and recipient, and edge weights denote compatibilities of the HLAs, which can be positive or negative. 
The network is \emph{indirectly observed}, as the edge weights are estimated from transplant outcomes rather than directly observed. 
We propose a latent space model for such indirectly-observed weighted and signed networks. 
We demonstrate that our latent space model can not only result in more accurate estimates of HLA compatibilities, but can also be incorporated into survival analysis models to improve accuracy for the downstream task of predicting graft survival times.
\end{abstract}


\maketitle

\section{Introduction}

Kidney transplantation is by far the preferred treatment for people suffering from end-stage renal disease (ESRD), an advanced state of chronic kidney disease. 
Despite the advantages of kidney transplants, most patients with ESRD are treated with dialysis, primarily because there exist an insufficient number of compatible donors for patients. 
Kidney transplants (and other organ transplants in general) inevitably fail over time, referred to as graft failure. 
These patients then require a replacement with another kidney transplant, or they must return to the waiting list while being treated with dialysis. 

The time to graft failure, or \emph{graft survival time}, is determined by a variety of factors. 
A significant biological factor affecting clinical survival times of transplanted organs is the compatibility between the \emph{human leukocyte antigens (HLAs)} of the organ donor and recipient. 
Mismatches between donor and recipient HLAs may cause the recipient's immune response to launch an attack against the transplanted organ, resulting in worse outcomes, including shorter graft survival times.
However, it is extremely rare to identify donors that have a perfect HLA match with recipients (less than 10\%), so most transplants (more than 90\%) involve different degrees of mismatched HLAs. 
Interestingly, HLA mismatches appear not to be equally harmful. 
Prior work indicates that some mismatches may still lead to good post-transplant outcomes \citep{doxiadis1996association, claas2004acceptable}, suggesting differing levels of compatibility between donor and recipient HLAs. 

Recently, \citet{nemati2021predicting} proposed an approach to encode donor and recipient HLAs into feature representations that accounts for biological mechanisms behind HLA compatibility. 
By adding these features to a Cox proportional hazards (CoxPH) model, they were able to improve the prediction accuracy for the graft survival time for kidney transplants. 
Their CoxPH model also provides estimates of the compatibilities between donor and recipient HLAs through the coefficients of the model. 
However, most donor-recipient HLA pairs are infrequently observed, with many HLA pairs occurring in less than 1\% of all transplants. 
These estimated compatibilities are thus extremely noisy, and in some cases, have standard errors as large or larger than the estimates themselves!

In this paper, we propose to represent HLA compatibility as a network, where the edge weight between a donor HLA and a recipient HLA denotes the compatibility. 
We propose a latent space model for the HLA compatibility network, which is a weighted, signed, and bipartite network with extremely noisy edge weights. 
The latent space model allows us to use compatibilities of other HLA pairs in the network to improve the estimated compatibility of a given HLA pair.

\begin{figure}[t]
\centering
\includegraphics[width=3.3in]{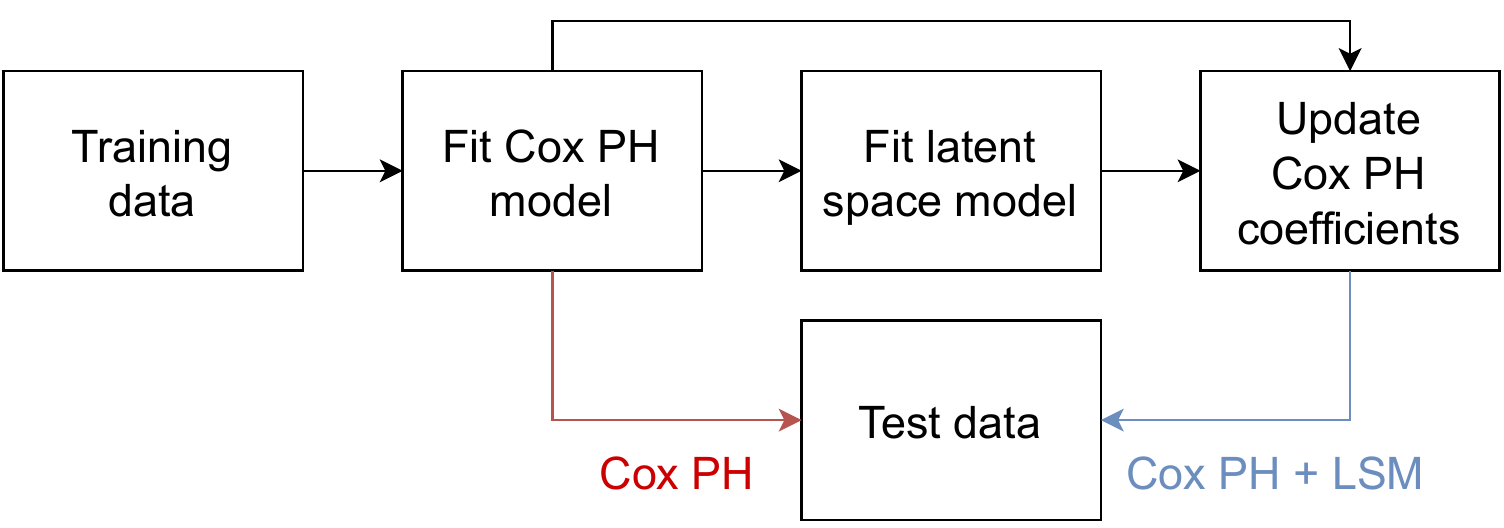}
\caption{End-to-end evaluation of HLA compatibility estimates using downstream task of survival prediction.
The HLA compatibility estimates from the Cox PH model (red line) are compared against the updated estimates from the latent space model (blue line) for survival prediction accuracy.}
\label{fig:CoxPH_with_LSM}
\end{figure}

Our main contributions are as follows:
\begin{itemize}
\item We introduce the notion of an \emph{HLA compatibility network} between donor and recipient HLAs, a noisy signed and weighted network estimated from outcomes of kidney transplants involving those HLAs.

\item We propose a latent space model for the HLA compatibility network to capture the underlying structure between HLAs and better predict their compatibilities.

\item We demonstrate that applying our latent space model to the HLA compatibility network results in more accurate predictions of the compatibilities.

\item We find that the predicted compatibilities from our latent space model can further improve accuracy for the downstream task (see Figure \ref{fig:CoxPH_with_LSM}) of predicting outcomes of kidney transplants, namely the graft survival times.

\end{itemize}

\section{Background}

\subsection{Human Leukocyte Antigens (HLAs)}
The HLA system is a system of proteins expressed on a transplant's cell that recognizes the immunogenicity of a kidney transplant from a donor \cite{claas2004acceptable,konvalinka2015utility}. The HLA system includes three primary HLA loci: HLA-A, HLA-B, and HLA-DR. For each HLA locus, there are many diverse HLA proteins. The general way to express HLA antigens is using two-digit, or low-resolution typing, where each HLA is identified by ``HLA'' followed by a hyphen: a letter indicating the locus, and a 1-digit or 2-digit number which indicates the HLA protein, e.g.~HLA-A1 \cite{shankarkumar2004human}. 
In this paper, we drop the ``HLA'' prefix and refer to HLA-A1 as simply A1.

Each person has two sets of HLAs, one inherited from the father and one from the mother. Therefore, each donor and recipient has 6 HLAs at these 3 loci, and anywhere from 0 to 6 of the HLAs can be mismatched.  
The compatibility between the mismatched HLAs of the kidney donor and recipient has shown to be a significant factor affecting transplanted kidneys' survival times \cite{claas2004acceptable,konvalinka2015utility,kosmoliaptsis2016alloantibody}. A kidney transplant recipient with higher HLA compatibility with the donor has a higher possibility of a good transplant outcome.

\subsection{Survival Analysis}
\label{sec:survival}
Survival analysis focuses on analyzing time-to-event data, where the objective is typically to model the expected duration until an event of interest occurs for a given subject \citep{wang2019machine}. 
For many subjects, however, the exact time of the event is unknown due to \emph{censoring}, which may occur for many reasons, including the event not yet occurring, or the subject dropping out from the study. 
We briefly introduce some survival analysis terminology and models relevant to this paper.

Let $T$ denote the time that the event of interest (graft failure) occurs. 
The hazard function is defined as 
\begin{equation*}
h(t) = \lim_{\Delta t \rightarrow 0} \frac{\Pr(t \leq T < t + \Delta t | T \geq t)}{\Delta t}
\end{equation*}
and denotes the rate of the event of interest occurring at time $t$ given that it did not occur prior to time $t$. 
A common assumption in survival analysis is the \emph{proportional hazards assumption}, which assumes that covariates are multiplicatively related to the hazard function. 
In the Cox proportional hazards (CoxPH) model, the hazard function takes the form
\begin{equation*}
h(t|\vec{x}_i) = h_0(t) \exp(\omega_0 + \omega_1 x_{i1} + \ldots + \omega_d x_{id}),
\end{equation*}
where $h_0(t)$ is a baseline hazard function, $x_{ij}$ denotes the $j$th covariate for  subject $i$, and $\omega_j$ denotes the coefficient for the $j$th covariate. 
Note that the hazard ratio for any two subjects $x_1, x_2$ is independent of the baseline hazard function $h_0(t)$. 
Assume that the $j$th covariate is binary.
If we consider two groups of subjects who differ only in the $j$th covariate, then the hazard ratio is given by $e^{\omega_j}$, and the log of the hazard ratio is thus $\omega_j$.

\subsection{Latent Space Models}
Network models are used to represent relations among interacting units. \citet{hoff2002latent} proposed a class of latent space models for networks, where the probability of a edge existence between two entities depends on their positions in an unobserved Euclidean space or latent space. Let $A$ denote the adjacency matrix of a network, with $a_{ij} = 1$ for node pairs $(i,j)$ with an edge and $a_{ij} = 0$ otherwise. The model assumes conditional independence between node pairs given the latent positions, and the log odds of an edge being formed between nodes $(i,j)$ is given by
$\alpha + \beta x_{i,j} - \|z_i - z_j\|$,
where $x_{ij}$ denotes observed covariates between nodes $(i,j)$, $z_i, z_j\in \mathbb{R}^d$ denotes the latent positions in a $d$-dimensional latent space for nodes $i$ and $j$, and $\alpha$ and $\beta$ are the linear parameters. Within this parameterization, two nodes have higher probability to form an edge if they have closer latent positions.

\subsection{Related Work}

\subsubsection{Extensions of Latent Space Models}
The latent space model provides a visual and interpretable spatial representation for relational data. 
The model of \citet{hoff2002latent} was extended by researchers for more complex network based on data structures, including bipartite networks \cite{friel2016interlocking}, discrete-time dynamic networks \cite{sewell2015latent}, and multimodal networks \cite{wang2019joint}. Node-specific random effects were added to the latent space model by \cite{hoff2005bilinear} to capture the degree heterogeneity. In this work, we propose a HLA latent space model for a signed, weighted, and bipartite indirectly-observed network to capture the relation within HLA compatibility networks.

\subsubsection{Predicting Kidney Transplant Outcomes}
A variety of approaches have been proposed for predicting outcomes for kidney transplants, as well as other organ transplants. 
Due to the high rate of censored subjects in this type of data, most approaches use some form of survival prediction, including an ensemble model that combines CoxPH models with random survival forests \cite{mark2019using} and a deep learning-based approach \cite{luck2017deep}.

The most relevant prior work is that of \citet{nemati2021predicting}, who propose a variety of feature representations for HLA for predicting graft survival time. 
They experimented with these different feature representations using CoxPH models, gradient boosted trees, and random survival forests and found that including HLA information could result in small improvements of up to 0.007 in the prediction accuracy as measured by Harrell's concordance index (C-index) \citep{harrell1982evaluating}. 

\section{HLA Compatibility Networks}

\subsection{Data Description}
\label{sec:data}
This study used data from the Scientific Registry of Transplant Recipients (SRTR). The SRTR data
system includes data on all donor, wait-listed candidates, and transplant recipients in the U.S., submitted by the
members of the Organ Procurement and Transplantation Network (OPTN). 
The Health Resources and Services
Administration (HRSA), U.S. Department of Health and Human Services provides oversight to the activities of the
OPTN and SRTR contractors.

We use the same inclusion criteria and data preprocessing as in \citet{nemati2021predicting}, which we briefly summarize in the following. 
We consider only transplants performed between the years 2000 and 2016 with deceased donors, recipients aged 18 years or older, and only candidates who are receiving their first transplant. 
We use \emph{death-censored graft failure} as the clinical endpoint (prediction target), so that patients who died with a functioning graft are treated as censored since they did not exhibit the event of interest (graft failure). 
For censored instances, the censoring date is defined to be the last follow-up date.
We consider a total of $106,372$ kidney transplants, for which $74.6\%$ are censored.

\subsubsection{HLA Representation}
We encode HLA information using the HLA types and pairs of the donor and recipient for each transplant. 
These HLA type and pair features are constructed according to the approach of \citet{nemati2021predicting}. 
HLA types for the donor and recipient are represented by a one-hot-like encoding, resulting in binary variables such as DON\_A1, DON\_A2, \ldots, REC\_A1, REC\_A2, \ldots, where the value for a donor (resp.~recipient) HLA type variable is one if the donor (resp.~recipient) possesses that HLA type. 
An HLA type that is a split of a broad type has ones for variables for both the split and broad.
For example, $\text{DON\_A23} = 1$ for a transplant if the donor possesses HLA type A23. 
Since A23 is a split of the broad A9, $\text{DON\_A9} = 1$ for this transplant also.

Donor-recipient HLA pairs are also represented by a one-hot-like encoding, resulting in binary variables such as DON\_A1\_REC\_A1, DON\_A1\_REC\_A2, \ldots, where the value for such an HLA pair variable is one if the donor and recipient possess the specified HLA types, and the HLA pair is active. 
(Some HLA pairs are inactive due to  asymmetry in the roles of donor and recipient HLAs; we refer interested readers to \citet{nemati2021predicting} for details.) 
Broads and splits are handled in the same manner as for HLA types. 
For example, $\text{DON\_A23\_REC\_A1} = 1$ for a transplant if the donor possesses A23, the recipient possess A1, and the HLA pair (A23, A1) is active. 
If this is the case, then $\text{DON\_A9\_REC\_A1} = 1$ also because A23 is a split of the broad A9.

\subsection{HLA Compatibility Network Construction}
While it is possible to construct a network of donors and recipients directly (i.e.~with donors and recipients as nodes and an edge denoting a transplant from a donor to a recipient), this directly-observed network is of very little interest scientifically. 
Each donor node has maximum degree of $2$ because that is the maximum number of kidneys they can donate. 
Each recipient node also has very small degree denoting the number of transplants they have received. 
The network consists of many small components that are not connected. 

We instead consider an indirectly-observed \emph{HLA compatibility network}, where nodes denote HLA types. 
We consider a separate set of donor nodes and recipient nodes since the effects of an HLA type may differ when it appears on the donor compared to recipient side. 
To avoid confusion between a donor and recipient node, we use uppercase letters for the HLA type of a donor node (e.g.~A1) and lowercase letters for the HLA types of a recipient node (e.g.~a1). 

We propose the following definition of HLA compatibility using hazard ratios, which are commonly used in survival analysis as described in Section \ref{sec:survival}. 
\begin{definition}
\label{def:compatibility}
We define the \emph{compatibility} between a donor HLA $d_i$ and a recipient HLA $r_j$ as sum of the negative log of the hazard ratios for donor $d_i$, for recipient $r_j$, and for the donor-recipient pair $(d_i, r_j)$.
\end{definition}
Let $\delta_i$ and $\gamma_j$ denote the negative logs of the hazard ratios for donor HLA type $d_i$ and recipient HLA type $r_j$, respectively. 
Let $\eta_{ij}$ denote the negative log of the hazard ratio for the donor-recipient HLA pair $(d_i, r_j)$. 
Then, the compatibility of donor HLA $d_i$ and recipient HLA $r_j$ from Definition \ref{def:compatibility} can be written as
\begin{equation}
\label{eq:compatibility}
\text{Compatibility}(d_i, r_j) = \delta_i + \gamma_j + \eta_{ij}.
\end{equation}

\subsubsection{Estimating HLA Compatibilities}
The compatibilities of the donor and recipient HLA types are unknown, so we must estimate them from the transplant data.
To estimate the compatibilities, we first remove all HLA types and pairs that were not observed in at least $100$ transplants\footnote{The estimated hazard ratios for these rarely occurring HLA types and pairs have extremely high standard errors, and in some cases, including them creates instabilities in estimating the CoxPH coefficients.}. 
We fit an $\ell_2$-penalized CoxPH model to the data. 
We use 2-fold cross-validation to tune the $\ell_2$ penalty parameter to maximize the partial log-likelihood on the validation folds. 
The result is a set of coefficients for each of the covariates, including basic covariates such as age and race, as well as the HLA types and the HLA pairs. 
The negated coefficients for the donor HLA type $d_i$, the recipient HLA type $r_j$, and the donor-recipient HLA pair $(d_i, r_j)$ can be substituted into \eqref{eq:compatibility} to obtain the estimated compatibility of $(d_i, r_j)$.  
Since positive coefficients in the CoxPH model are associated with higher probability of graft failure, we negate the estimated coefficients so that positive compatibilities are associated with lower probability of graft failure, i.e.~better transplant outcomes. 

\subsubsection{HLA Compatibility Network}
\label{sec:hla_network}
The HLA compatibilities can be represented as a network with both node and edge weights. 
There are two types of nodes in the network: donor nodes and recipient nodes. 
Each donor node has an unknown true weight $\delta_i$, and each recipient node has an unknown true weight $\gamma_j$. 
Each edge connects a donor node $d_i$ to a recipient node $r_j$ and has true weight $\eta_{ij}$. 

In the HLA compatibility network, the true node and edge weights are terms from \eqref{eq:compatibility} and are unobserved. 
We observe only a noisy version of the weights in the form of the estimated CoxPH coefficients. 
We model the observed node and edge weights as independent realizations of Gaussian random variables in the following manner:
\begin{itemize}
\item Observed donor node weight: $y_{d_i} \sim \mathcal{N}(\delta_i, \sigma_{\delta_i}^2)$.
\item Observed recipient node weight: $y_{r_j} \sim \mathcal{N}(\gamma_j, \sigma_{\gamma_j}^2)$.
\item Observed edge weight by $w_{ij} \sim \mathcal{N}(\eta_{ij}, \sigma_{ij}^2)$.
\end{itemize}
We can thus view the observed node and edge weights as estimates of the true node and edge weights, respectively. 
The estimated CoxPH coefficients also have estimated standard errors, which we can use as estimates for $\sigma_{\delta_i}, \sigma_{\gamma_j}, \sigma_{ij}$. 

\begin{figure}[t]
    \centering
    \includegraphics[width=3.2in]{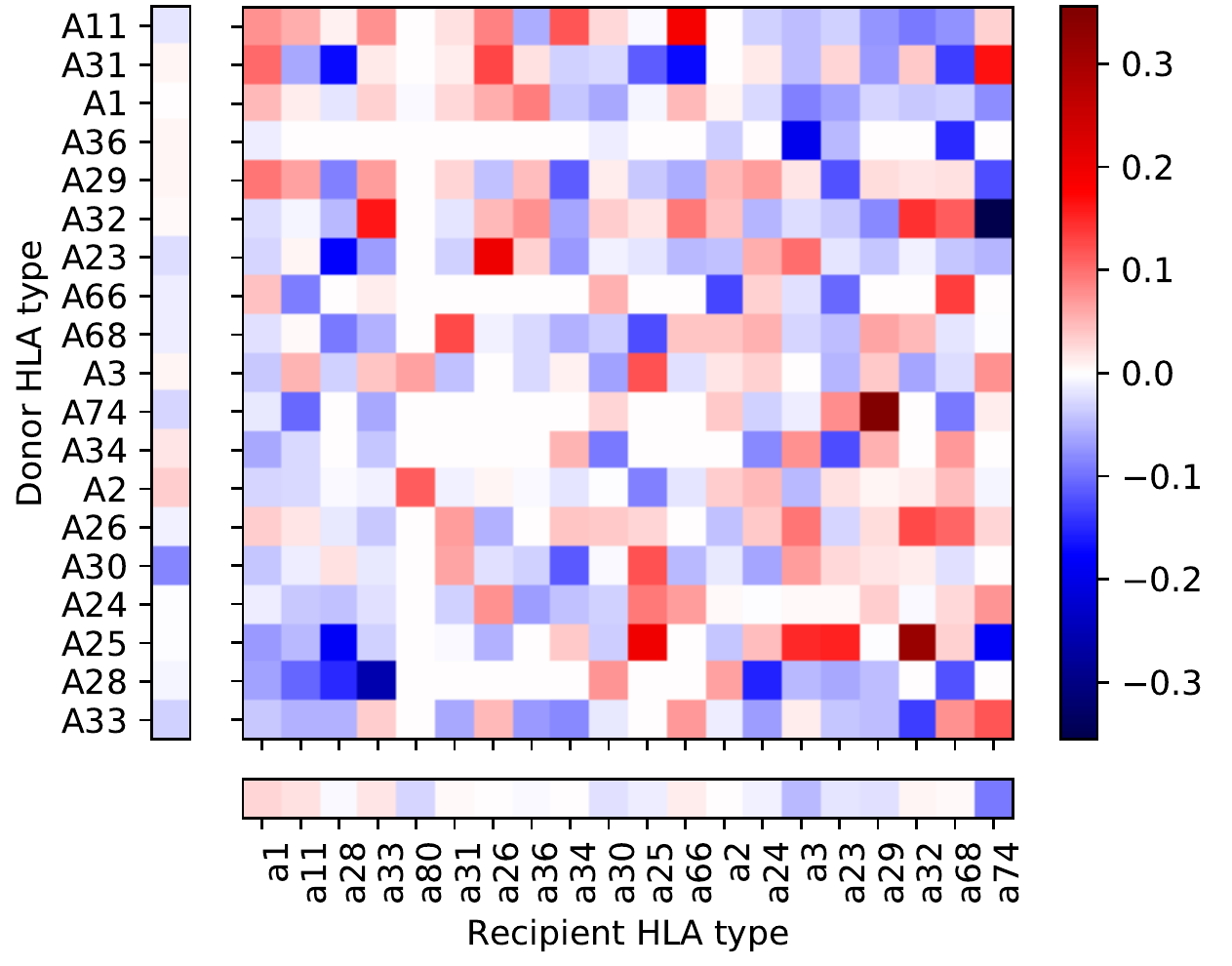}
    \caption{Heat maps illustrating node and edge weights in the HLA compatibility networks for HLA-A. 
    Rows and columns are ordered by their positions from fitting a 1-D latent space model to the network. 
    Entries along the diagonal should be more positive (red), as these nodes are closer in the latent space, and entries should turn more negative (blue) further away from the diagonal.}
    \label{fig:heatmaps}
\end{figure}

There are 3 separate HLA compatibility networks, one for each of the 3 loci: HLA-A, HLA-B, and HLA-DR. 
The HLA-A compatibility network estimated using the observed node and edge weights is shown in Figure \ref{fig:heatmaps}. 
Notice that node and edge weights can be both positive and negative. 
All donor-recipient pairs that have been observed in at least $100$ transplants contain an edge in the network. 
The variation in edge weights can be quite large, ranging from roughly $-0.3$ to $0.3$ in A and DR and about $-0.4$ to $0.4$ in B. 

\section{HLA Latent Space Model}
We utilize a latent space model to learn the hidden features underlying the HLA compatibility network. Within the latent space model, donors' and recipients' node positions are embedded in the latent space, where a donor and recipient in the observed HLA network with higher edge weight are put closer. Thus, the HLA compatibilities could be induced through distances between nodes in the latent space.

\subsection{Model Description}
Let $z_{d_i}$ and $z_{r_j}$ denote the positions of the $i$th donor node and the $j$th recipient node, respectively, in the latent space. Let $N_d$ and $N_r$ denote the number of donor nodes and recipient nodes, respectively. 
Unlike the latent distance model of \citet{hoff2002latent}, which models a binary relation between two nodes using a logistic regression, we model the affinity (true edge weight) between a donor and recipient node using the linear relation $\eta_{ij} = \alpha - \beta \, ||z_{d_i} - z_{r_j}||^2$. 
Moreover, we employ donor and recipient node effect terms $\delta_i, \gamma_j$ to the model as in \cite{hoff2005bilinear, wang2019joint} to capture the true node weights. 
Then, the compatibility between donor node $d_i$ and recipient node $r_j$ is given by
\begin{equation}
\label{eq:true_edge_weight}
    \mu_{ij} = \eta_{ij} + \delta_i + \gamma_j = \alpha - \beta \, ||z_{d_i} - z_{r_j}||^2 + \delta_i + \gamma_j,
\end{equation}
where $\alpha$ and $\beta$ are scalar parameters.
The observed data consists of the observed node and edge weights $\{y_{d_i}, y_{r_j}, w_{ij}\}$, modeled as described in Section \ref{sec:hla_network}.

\subsection{Estimation Procedure}
The proposed model has the set of unknown parameters $\theta = (Z_d, Z_r, \alpha, \beta, \gamma, \delta)$. $z_{d_i} \in Z_d$ and $z_{r_i} \in Z_r$ represent the latent positions of donor and recipient nodes, respectively, in a $d$-dimensional Euclidean space. Therefore, $Z_d$ and $Z_r$ are $N_d \times d$ and $N_r \times d$ matrices, respectively. Each pair of nodes is associated with a donor node effect $\gamma_i \in \gamma$ and a recipient node effect $\delta_i \in \delta$. $\alpha$ and $\beta$ are the slope and intercept terms of the linear relationship. We constrain the slope $\beta$ to be positive to keep the node pairs with higher edge weights closer together in the latent space.

Beginning with the likelihood of the latent distance model derived by \citet{hoff2002latent}, we can write the likelihood of our proposed HLA latent space model as
\begin{equation*}
\label{eq: logll}
\begin{split}
    p(W, y_d, y_r|\theta) = \prod_{i=1}^{n_1}\prod_{j=1}^{n_2} p&(w_{ij}|z_d, z_r, \alpha, \beta) \\
    &\prod_{i=1}^{n_1}p(y_{d_i}|\delta_i)  \prod_{j=1}^{n_2}p(y_{r_j}|\gamma_j),
\end{split}
\end{equation*}
where $\theta$ denotes the set of all unknown parameters.
The probability distributions for the observations are given by
\begin{align*}
    p(w_{ij}|z_d, z_r, \alpha, \beta) &= \mathcal{N}(w_{ij}|\eta_{ij}, \sigma_{ij}^2) \\
    p(y_{d_i}|\delta_i) &= \mathcal{N}(y_{d_i}|\delta_i, \sigma^2_{\delta_i}) \\
    p(y_{r_j}|\gamma_j) &= \mathcal{N}(y_{r_j}|\gamma_j, \sigma^2_{\gamma_j}),
\end{align*}
where $\sigma_{ij}^2$ denotes the variance of the observed edge weights, and $\sigma^2_{\delta_i}, \sigma^2_{\gamma_j}$ denote the variances of the observed node weights. 
We use plug-in estimators for these variance parameters using the estimated standard errors for the CoxPH coefficients.

We optimize the log-likelihood of the model parameters using the L-BFGS-B \citep{byrd1995limited} optimizer implemented in SciPy.

\paragraph*{Parameter Initialization}
We employ a multidimensional scaling (MDS) algorithm \cite{cox2008multidimensional} as an initialization for the latent node positions. MDS attempts to find a set of positions where each point represents one of the entities, and the distances between points depend on their dissimilarities for each pair of entities. As the HLA compatibility network is a weighted bipartite network, we do not have edge weights between node pairs within the same group: donor-donor edge weights and recipient-recipient edge weights. Instead, we use the correlation coefficients between nodes based on their edge weights with the other type of node. 
For example, we compute the correlation coefficient between the edge weights of two donors nodes with all recipient nodes. 
We then define the dissimilarity matrix with entries given by
$d_{ij} = 1 - \text{logistic}(w_{ij})$,
where $w_{ij}$ represents the weights or correlation coefficients between node pair $(i,j)$, depending on whether they are nodes of different types or the same type, respectively. All other parameters are initialized randomly.

\section{Simulation Experiments}

To make a pilot evaluation of our proposed model, we fit our model to simulated networks. We simulate the HLA bipartite networks with number of nodes $N_d = 20, N_r = 20$, parameters $\alpha = 1, \beta = 1$, and latent dimension $d = 2$. The latent positions are sampled independently from a 2-D normal distribution: $z_d, z_r \sim \mathcal{N}(0,0.5I)$. Donor and recipient effects are sampled independently from a 1-D Normal distribution: $\delta_d, \gamma_r \sim \mathcal{N}(0,0.5)$. We compute the true weights using \eqref{eq:true_edge_weight}. The noise of the edge weights is controlled by a scalar $\sigma_w^2$. 

Similar to the latent space model of \citet{huang2022mutually}, $\beta$ is not identifiable because it enters multiplicatively into \eqref{eq:true_edge_weight}. 
The latent positions $Z$ are also not identifiable in a latent space model, as noted by \citet{hoff2002latent}, and can only be identified up to a rotation. 
Thus, we set $\beta = 1$ in all simulations and use a Procrustes transform to rescale and rotate the latent positions to best match the true latent positions.

\paragraph*{Low noise simulated network}
We first fit the HLA latent space model to the low noise simulated network by setting $\sigma_w = 0.15$. 
After fitting the model, we compute the root mean squared error (RMSE) and $R^2$ values between the actual parameters and estimated parameters, which are shown in Table \ref{tab: sim_1}. Both RMSE and $R^2$ indicate an accurate prediction for all the parameters and the edge weights.

\paragraph*{High noise simulated network}
Since the real kidney transplant data has extremely noisy edge weights, we conduct another experiment by setting a high variance to the simulation where $\sigma_w$ = $1.5$. As we increase the noise, the RMSE increases as expected. The $R^2$ for nodal effect $\delta$ and $\gamma$ still indicate reasonable estimates. The $R^2$ for $w$, $z_d$, and $z_r$ is about 0.5, indicating moderately accurate estimates, which we consider acceptable for a high noise network.

\begin{table}[t]
\caption{Error between estimated and actual parameters by fitting the HLA latent space model to 15 simulated networks. Mean $\pm$ standard error is shown for each parameter.}
\label{tab: sim_1}
\centering   
\begin{tabular}{ccccc}
\hline
    &    &Parameters  & Low noise   & High noise \\
\hline
\multirow{6}{*}{RMSE}  &     &$w$ & $0.124\pm 0.001 $  &$1.264\pm 0.014 $\\
    &    &$z_d$       & $0.014\pm 0.002$ &$0.111\pm 0.007 $\\
    &    &$z_r$       & $0.024\pm 0.010 $ &$0.101\pm 0.006 $\\
    &    &$\delta$    & $0.105\pm 0.018 $  &$0.173\pm 0.015 $\\
    &    &$\gamma$    & $0.097\pm 0.015 $  &$0.170\pm 0.012 $\\
    &    &$\alpha$    & $0.047\pm 0.068 $  &$0.779\pm 0.107 $      \\
\hline
\multirow{5}{*}{$R^2$}  
    &    &$w$       & $0.990\pm 0.001 $ &$0.555\pm 0.018 $\\
    &    &$z_d$       & $0.989\pm 0.004 $ &$0.482\pm 0.064 $\\
    &    &$z_r$    & $0.989\pm 0.004 $  &$0.571\pm 0.056 $\\
    &    &$\delta$    & $0.937\pm 0.026 $  &$0.860\pm 0.024 $\\
    &    &$\gamma$    & $0.924\pm 0.034 $  &$0.854\pm 0.025 $      \\    
\hline
\end{tabular}
\end{table}

\section{Real Data Experiments}

\subsection{Model-based Exploratory Analysis}
\label{sec:exploratory}

\begin{figure*}[tp]
    \newcommand{\figwidth}{3.4in}
    \centering
    \hfill
    \begin{subfigure}[c]{\figwidth}
        \centering
        \includegraphics[width=\figwidth]{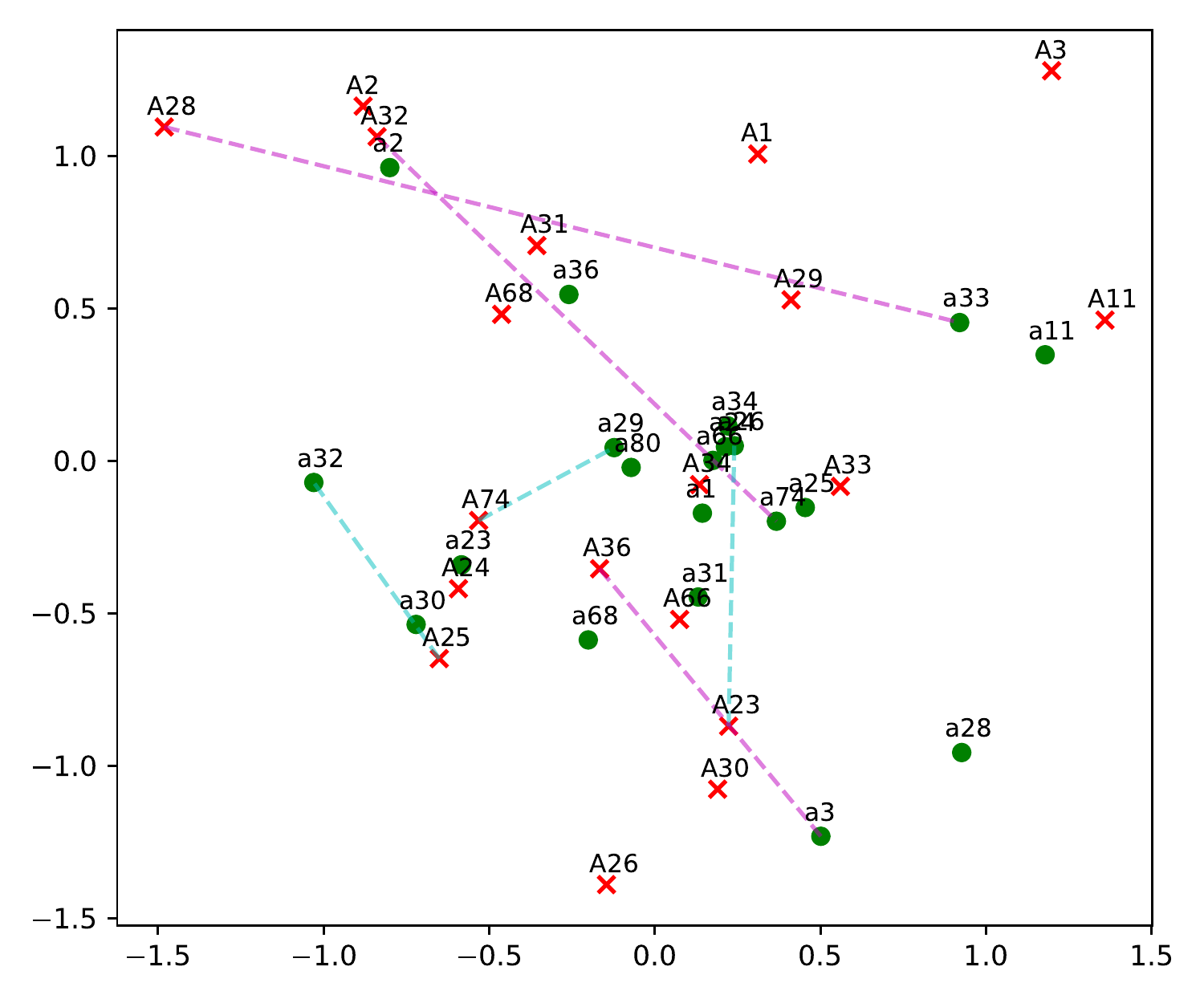}
        \caption{HLA-A}
        \label{fig: hla_a}
    \end{subfigure}
    \hfill
    \begin{subfigure}[c]{\figwidth}
        \centering
        \includegraphics[width=\figwidth]{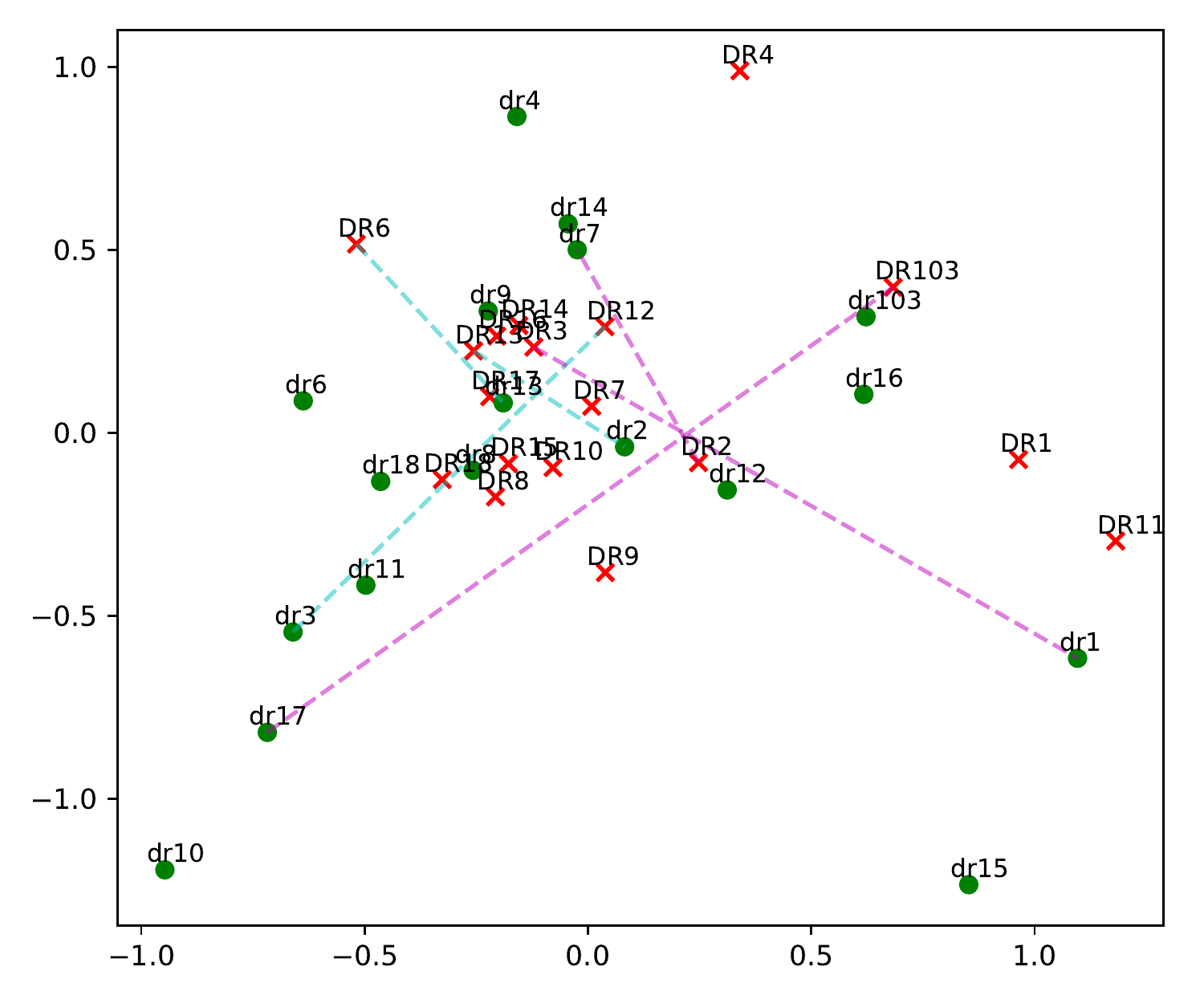}
        \caption{HLA-DR}
        \label{fig: hla_dr}
    \end{subfigure}
    \hfill
    \\[6pt]
    \renewcommand{\figwidth}{5.4in}
    \begin{subfigure}[c]{\figwidth}
        \centering
        \includegraphics[width=\figwidth]{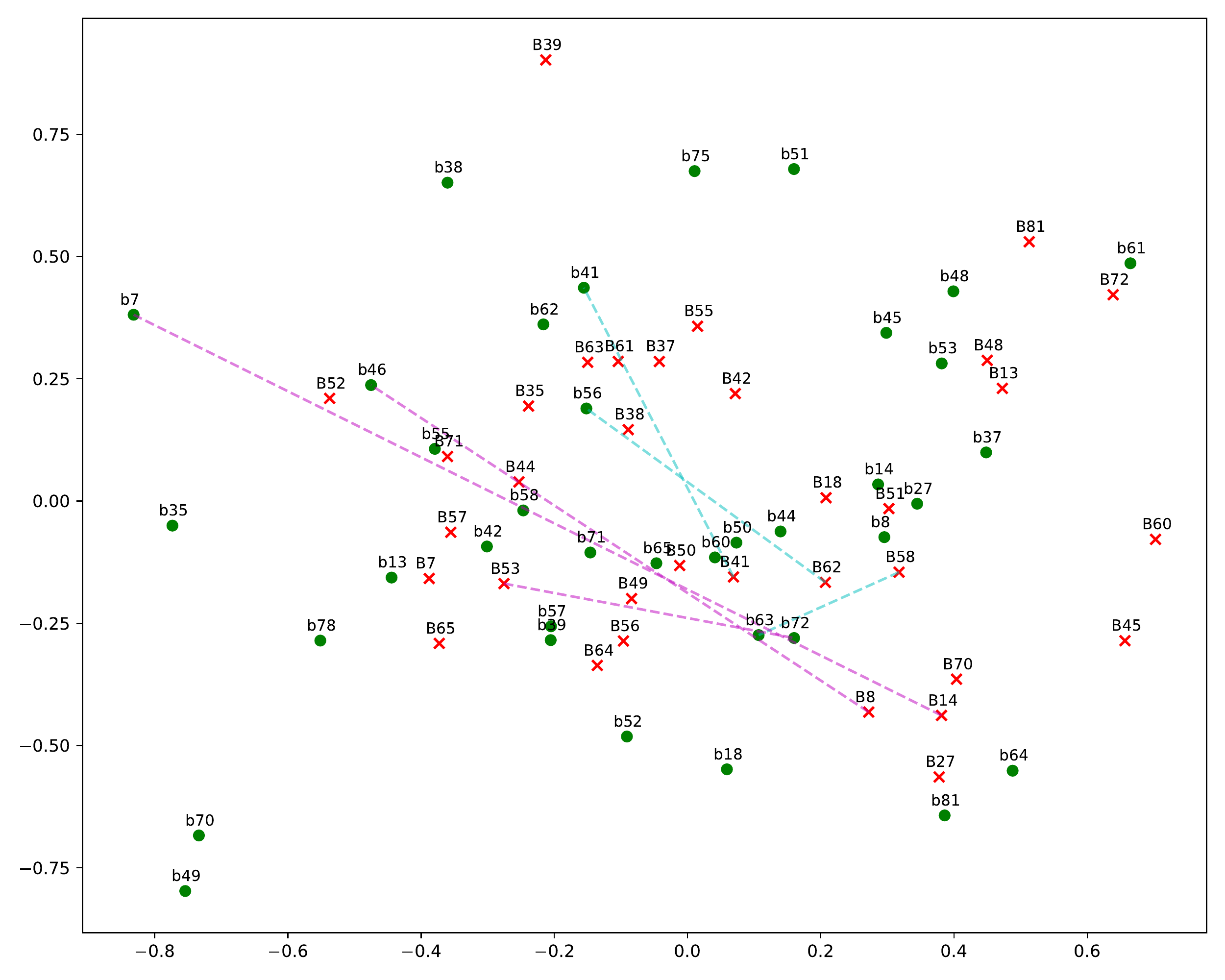}
        \caption{HLA-B}
        \label{fig: hla_b}
    \end{subfigure}
    \hfill
    \begin{subfigure}[c]{1.6in}
        \renewcommand{\tabcolsep}{3pt}
        \centering
        \footnotesize
        \begin{tabular}{cc}
        \hline
        Top 3 highest & Weight $\pm$ s.e. \\
        \hline
        (A74, a29)    & $0.349 \pm 0.171$ \\
        (A25, a32)    & $0.319 \pm 0.139$ \\
        (A23, a26)    & $0.198 \pm 0.103$ \\
        \hline
        (B58, b63)    & $0.384 \pm 0.167$ \\
        (B41, b41)    & $0.326 \pm 0.160$ \\
        (B62, b56)    & $0.305 \pm 0.174$ \\
        \hline
        (DR12, dr3)   & $0.389 \pm 0.164$ \\
        (DR13, dr2)   & $0.271 \pm 0.151$ \\
        (DR6, dr13)   & $0.230 \pm 0.135$ \\
        \hline 
        \\
        \hline
        Top 3 lowest & Weight $\pm$ s.e.\\
        \hline                    
        (A32, a74)   & $-0.356 \pm 0.096$\\
        (A28, a33)   & $-0.257 \pm 0.141$\\
        (A36, a3)    & $-0.197 \pm 0.136$\\
        \hline
        (B53, b72)   & $-0.472 \pm 0.129$\\
        (B8, b46)    & $-0.425 \pm 0.158$\\
        (B14, b7)    & $-0.279 \pm 0.105$\\
        \hline
        (DR103, dr17)& $-0.253 \pm 0.120$\\
        (DR3, dr1)   & $-0.235 \pm 0.111$\\
        (DR2, dr7)   & $-0.230 \pm 0.152$\\
        \hline 
        \end{tabular}
        \caption{Donor-recipient HLA pairs with 3 highest and 3 lowest edge weights ($\pm$ standard errors)}
        \label{tab: sanity}
    \end{subfigure}
    \hfill
    \caption{The 2-D latent space plot for HLA networks for all 3 loci. 
    Red x: donor node; green circle: recipient node.
    The 3 highest and lowest edge weights are shown with dashed cyan and magenta lines, respectively. 
    Donor-recipient pairs with the highest edge weights tend to be placed closer together in the latent space compared to those with the lowest edge weights.}
    \label{fig: hla}
\end{figure*}

In the HLA latent space model, we expect negative relationships between HLA pair coefficients (edge weights) and distances in Euclidean space. We fix the latent dimension to be $d=2$ in three HLA networks (A, B, DR) to visualize the latent positions of the nodes. After fitting the HLA latent space model, the 2-D latent space plots we obtained for all three HLA loci are shown in Figure \ref{fig: hla_a}-\subref{fig: hla_b}. 

From examining the latent positions, we find that pairs of nodes with higher edge weights tend to appear closer together in the latent space and vice versa. For example, in the HLA-A plot, donor A74 and recipient a29 have a high weight of $0.349$ and tend to be placed close together in the 2-D plot. Conversely, donor A28 and recipient a33 have a low weight of $-0.257$ and are placed on opposite sides of the latent space. 
We draw cyan dashed lines indicating the top 3 highest weight pairs and magenta dashed lines indicating the top 3 lowest weight pairs in all three 2-D HLA latent space plots. 
These weights and standard errors are also shown in Figure \ref{tab: sanity}. 
By comparing the latent space plot and the edge weights, we discover that A and B have a relatively good fit, while DR is slightly worse, with some cyan lines longer than magenta. 

\subsection{HLA Compatibility Prediction}
\label{sec:hla_pred_exp}

Next, we evaluate the ability of our latent space model to predict HLA compatibilities.
We split the transplant data 50/50 into training/test, and we fit the $\ell_2$-penalized CoxPH model two times, once each on the training and test folds.
We obtained two sets of HLA compatibility data: $\HLA_1$ denotes the first (training) fold, and $\HLA_2$ denotes the second (test) fold.
We then fit the proposed HLA latent space model on the $\HLA_1$ data for varying latent dimension $d$ and evaluate prediction accuracy on the $\HLA_2$ data.

\subsubsection{Evaluation Metrics}
We consider 3 measures of prediction accuracy:
\begin{itemize}
\item Root mean squared error (RMSE) between the predicted and observed compatibilities on $\HLA_2$. 
\item Mean log-probability on $\HLA_2$: We use a Gaussian distribution with mean given by the observed compatibilities on $\HLA_2$ and standard deviation given by its standard error to compute the log probability. This is equivalent to a negated weighted RMSE where higher weight is given to HLA pair coefficients with smaller standard errors.
\item Sign prediction accuracy: We threshold the predicted and observed compatibilities at 0 and compute the binary classification accuracy. 
\end{itemize}
We consider also the prediction accuracy for the downstream task of graft survival time prediction, which we describe in Section \ref{sec:survival_pred}.

\subsubsection{Comparison Baselines}
We compare our proposed latent space model against 3 other methods:
\begin{itemize}
    \item Cox proportional hazards (CoxPH): Directly uses the estimated CoxPH coefficients without any further processing.
    \item Non-negative matrix tri-factorization (NMTF) \cite{li2009non}: A technique for learning low-dimensional feature representation of relational data that decomposes the given matrix into three smaller matrices rather than two matrices as in standard non-negative matrix factorization. As we have negative values in our HLA network, we first apply a logistic function to transform all compatibilities to $(0, 1)$. We then apply NMTF. Finally, we use a logit function to transform the values back to the original domain with both positive and negative values.
    \item Principal component analysis (PCA): A classical linear dimensionality reduction technique that projects the given matrix into a lower dimensional space. 
    We reconstruct the compatibilities from the first $d$ principal components.
\end{itemize}
For all of the methods, we choose the number of dimensions $d$ that returns the best mean log-probability on $\HLA_2$.

\subsubsection{Results}
\begin{table}[t]
\caption{Evaluation metrics for HLA compatibility and graft survival time prediction. 
Our proposed LSM performs competitively on HLA compatibility prediction on all 3 loci and achieves the best C-index for graft survival time prediction.
}
\label{tab: hla_rmse}
\centering   
\begin{tabular}{ccccc}
\hline
Metric          & Model        & HLA-A  & HLA-B  & HLA-DR \\
\hline
\multirow{4}{*}{RMSE}            & CoxPH        & 0.159 & 0.191 & 0.168 \\
                & CoxPH + LSM  & \bf 0.110 &  \bf 0.139 &  0.129 \\
                & CoxPH + NMTF & 0.112 &  0.140 & \bf 0.122 \\
                & CoxPH + PCA  & 0.124 &  0.149 &  0.138 \\
\hline
\multirow{4}{*}{Mean log-prob.}  & CoxPH        & 0.609 & 0.374 & 0.541 \\
        & CoxPH + LSM          &  0.911 &  0.657  &  0.800 \\
        & CoxPH + NMTF         & \bf 0.936 & \bf 0.659  & \bf 0.885 \\
        & CoxPH + PCA          &  0.887 &  0.606  &  0.776 \\
\hline
\multirow{4}{*}{Sign prediction}  & CoxPH       & 0.535 & 0.515 & 0.510\\
        &CoxPH + LSM           & \bf 0.542 & \bf 0.576  &  0.552 \\
        &CoxPH + NMTF          & \bf 0.542 & 0.522  & \bf 0.573 \\
        &CoxPH + PCA           & 0.535 & 0.528  &   0.564 \\
\hline
\multirow{4}{*}{C-index}         & CoxPH        & \multicolumn{3}{c}{0.614} \\
                & CoxPH + LSM  & \multicolumn{3}{c}{\bf 0.625} \\
                & CoxPH + NMTF & \multicolumn{3}{c}{0.623} \\
                & CoxPH + PCA  & \multicolumn{3}{c}{0.621} \\
\hline
\end{tabular}
\end{table}

From Table \ref{tab: hla_rmse}, note that the predicted weights using other models to refine the estimated compatibilities are more accurate for each of the 4 metrics and 3 loci compared to directly using the CoxPH coefficients. Among the refinement methods, the LSM and NMTF have similar prediction accuracy, and both significantly outperform PCA. 
While the NMTF is competitive to our LSM in prediction accuracy, the LSM can also provide useful interpretations through the latent space, as shown in Section \ref{sec:exploratory}. 
Finally, notice the difficulty of the HLA compatibility prediction problem---the sign prediction accuracy on all 3 loci are quite low, from $51\%$ to $58\%$, despite the improvement from using the refinement.

\subsection{Graft Survival Time Prediction}
\label{sec:survival_pred}
One difficulty of evaluating the HLA compatibility predictions from Section \ref{sec:hla_pred_exp} is that the CoxPH coefficients from both data splits have very high standard errors. 
As a result, the prediction target (the HLA compatibility on the test set) is extremely noisy. 
Recall that the initial HLA compatibility estimates are obtained from fitting a CoxPH model tuned to maximize prediction accuracy for the graft survival times.
It is unclear whether improved prediction accuracy of the HLA compatibilities on the test set also lead to improved prediction accuracy of graft survival.

We thus propose an \emph{end-to-end} evaluation of our HLA compatibility estimates by incorporating them into the graft survival prediction, as shown in Figure \ref{fig:CoxPH_with_LSM}. 
After fitting the latent space model, we replace the CoxPH coefficients for the HLA types with the negated estimates of donor and recipient effects given by $-\hat{\gamma_i}$ and $-\hat{\delta_j}$, respectively. 
We also replace the CoxPH coefficients for the HLA pairs with the negated estimates of the donor-recipient pair weights $-\hat{\eta}_{ij}$.

The question we are seeking to answer here is as follows: \emph{Does replacing the estimated CoxPH coefficients with the estimated HLA compatibilities from our latent space model improve accuracy on the downstream task---graft survival prediction?} 
To evaluate this, we use the same 50/50 training/test splits from Section \ref{sec:hla_pred_exp} and compare the C-indices on the test splits between the trained CoxPH model and the updated CoxPH coefficients using our latent space model.

The answer to this question is yes, as shown by the C-indices in Table \ref{tab: hla_rmse}. 
Notice that our proposed latent space model improves the C-index by about 0.011 compared to directly using the CoxPH coefficients. 
The NMTF and PCA provide smaller improvements of 0.009 and 0.007, respectively. 
We note that an improvement of 0.011 in the C-index for graft survival prediction in kidney transplantation is a large improvement! 
For comparison, using the same dataset and inclusion criteria, \citet{nemati2021predicting} evaluated C-indices using 6 different HLA representations across 3 different survival prediction algorithms and found achieved a maximum improvement of 0.007. 
Similarly, using the same dataset but slightly different inclusion criteria, \citet{luck2017deep} achieved a maximum improvement of 0.005.
The appreciable improvement in C-index demonstrates the utility of our latent space modeling approach not only for interpreting HLA compatibilities, but also for graft survival prediction. 

\section{Conclusion}
We proposed a model for HLA compatibility in kidney transplantation using an indirectly-observed weighted and signed network.  
The weights were estimated from a CoxPH model fit to data on over $100,000$ transplants, yet they are very noisy, with standard errors sometimes on the same order of magnitude as the mean weight estimates themselves.

Our main contribution was to develop a latent space model for the HLA compatibility network. 
The latent space model provided both an interpretable visualization of the HLA compatibilities and used the network structure to improve the estimated compatibilities. 
We found that the latent space model not only resulted in more accurate estimated compatibilities, but also improved graft survival prediction accuracy when the estimates were substituted back into the CoxPH model as coefficients.

\paragraph*{Limitations and Future Work}
We chose to use a linear model for the HLA compatibilities for simplicity. 
Both compatibility estimation and survival prediction accuracy could also potentially be improved by incorporating non-linearities into the model. 
Furthermore, if we consider improving survival prediction accuracy as our objective, then our proposed approach can be viewed as a two-stage process: first estimate weights by maximizing the CoxPH partial log-likelihood, and then refine the weight estimates by maximizing the latent space model log-likelihood. 
The improvement in survival prediction accuracy from this two-stage process suggests that accuracy could potentially be further improved by jointly maximizing both objective functions. 
Indeed, this is an interesting avenue for future work that we are currently exploring.

\section*{Acknowledgment}
The authors thank Robert Warton, Dulat Bekbolsynov, and Stanislaw Stepkowski for their assistance with the HLA data.

The research reported in this publication was supported by the National Library of Medicine of the National Institutes of Health under Award Number R01LM013311 as part of the NSF/NLM Generalizable Data Science Methods for Biomedical Research Program. The content is solely the responsibility of the authors and does not necessarily represent the official views of the National Institutes of Health.

The data reported here have been supplied by the Hennepin Healthcare Research Institute (HHRI) as the contractor for the Scientific Registry of Transplant Recipients (SRTR). The interpretation and reporting of these data are the responsibility of the author(s) and in no way should be seen as an official policy of or interpretation by the SRTR or the U.S.~Government.
Notably, the principles of the Helsinki Declaration were followed.

\bibliographystyle{IEEEtranN}
\bibliography{references}

\begin{thebibliography}{19}
\providecommand{\natexlab}[1]{#1}
\providecommand{\url}[1]{#1}
\csname url@samestyle\endcsname
\providecommand{\newblock}{\relax}
\providecommand{\bibinfo}[2]{#2}
\providecommand{\BIBentrySTDinterwordspacing}{\spaceskip=0pt\relax}
\providecommand{\BIBentryALTinterwordstretchfactor}{4}
\providecommand{\BIBentryALTinterwordspacing}{\spaceskip=\fontdimen2\font plus
\BIBentryALTinterwordstretchfactor\fontdimen3\font minus
  \fontdimen4\font\relax}
\providecommand{\BIBforeignlanguage}[2]{{%
\expandafter\ifx\csname l@#1\endcsname\relax
\typeout{** WARNING: IEEEtranN.bst: No hyphenation pattern has been}%
\typeout{** loaded for the language `#1'. Using the pattern for}%
\typeout{** the default language instead.}%
\else
\language=\csname l@#1\endcsname
\fi
#2}}
\providecommand{\BIBdecl}{\relax}
\BIBdecl

\bibitem[Doxiadis et~al.(1996)Doxiadis, Smits, Schreuder, Persijn, van
  Houwelingen, van Rood, and Claas]{doxiadis1996association}
I.~I.~N. Doxiadis, J.~M.~A. Smits, G.~M.~{\relax Th}. Schreuder, G.~G. Persijn,
  H.~C. van Houwelingen, J.~J. van Rood, and F.~H.~J. Claas, ``Association
  between specific {HLA} combinations and probability of kidney allograft loss:
  the taboo concept,'' \emph{The Lancet}, vol. 348, no. 9031, pp. 850--853,
  1996.

\bibitem[Claas et~al.(2004)Claas, Witvliet, Duquesnoy, Persijn, and
  Doxiadis]{claas2004acceptable}
F.~H.~J. Claas, M.~D. Witvliet, R.~J. Duquesnoy, G.~G. Persijn, and I.~I.~N.
  Doxiadis, ``The acceptable mismatch program as a fast tool for highly
  sensitized patients awaiting a cadaveric kidney transplantation: short
  waiting time and excellent graft outcome,'' \emph{Transplantation}, vol.~78,
  no.~2, pp. 190--193, 2004.

\bibitem[Nemati et~al.(2021)Nemati, Zhang, Sloma, Bekbolsynov, Wang,
  Stepkowski, and Xu]{nemati2021predicting}
M.~Nemati, H.~Zhang, M.~Sloma, D.~Bekbolsynov, H.~Wang, S.~Stepkowski, and
  K.~S. Xu, ``Predicting kidney transplant survival using multiple feature
  representations for {HLA}s,'' in \emph{Proc. 19th Int. Conf. Artif. Intell.
  Med.}, 2021, pp. 51--60.

\bibitem[Konvalinka and Tinckam(2015)]{konvalinka2015utility}
A.~Konvalinka and K.~Tinckam, ``Utility of {HLA} antibody testing in kidney
  transplantation,'' \emph{J. Am. Soc. Nephrol.}, vol.~26, no.~7, pp.
  1489--1502, 2015.

\bibitem[Shankarkumar(2004)]{shankarkumar2004human}
U.~Shankarkumar, ``The human leukocyte antigen ({HLA}) system,'' \emph{Int. J.
  Hum. Genet.}, vol.~4, no.~2, pp. 91--103, 2004.

\bibitem[Kosmoliaptsis et~al.(2016)Kosmoliaptsis, Mallon, Chen, Bolton,
  Bradley, and Taylor]{kosmoliaptsis2016alloantibody}
V.~Kosmoliaptsis, D.~Mallon, Y.~Chen, E.~M. Bolton, J.~A. Bradley, and C.~J.
  Taylor, ``Alloantibody responses after renal transplant failure can be better
  predicted by donor--recipient {HLA} amino acid sequence and physicochemical
  disparities than conventional {HLA} matching,'' \emph{Am. J. Transplant.},
  vol.~16, no.~7, pp. 2139--2147, 2016.

\bibitem[Wang et~al.(2019{\natexlab{a}})Wang, Li, and Reddy]{wang2019machine}
P.~Wang, Y.~Li, and C.~K. Reddy, ``Machine learning for survival analysis: A
  survey,'' \emph{ACM Comput. Surv.}, vol.~51, no.~6, pp. 1--36, 2019.

\bibitem[Hoff et~al.(2002)Hoff, Raftery, and Handcock]{hoff2002latent}
P.~D. Hoff, A.~E. Raftery, and M.~S. Handcock, ``Latent space approaches to
  social network analysis,'' \emph{J. Am. Stat. Assoc.}, vol.~97, no. 460, pp.
  1090--1098, 2002.

\bibitem[Friel et~al.(2016)Friel, Rastelli, Wyse, and
  Raftery]{friel2016interlocking}
N.~Friel, R.~Rastelli, J.~Wyse, and A.~E. Raftery, ``Interlocking directorates
  in irish companies using a latent space model for bipartite networks,''
  \emph{Proc. Natl. Acad. Sci.}, vol. 113, no.~24, pp. 6629--6634, 2016.

\bibitem[Sewell and Chen(2015)]{sewell2015latent}
D.~K. Sewell and Y.~Chen, ``Latent space models for dynamic networks,''
  \emph{J. Am. Stat. Assoc.}, vol. 110, no. 512, pp. 1646--1657, 2015.

\bibitem[Wang et~al.(2019{\natexlab{b}})Wang, Paul, and
  De~Boeck]{wang2019joint}
S.~S. Wang, S.~Paul, and P.~De~Boeck, ``Joint latent space model for social
  networks with multivariate attributes,'' \emph{arXiv preprint
  arXiv:1910.12128}, 2019.

\bibitem[Hoff(2005)]{hoff2005bilinear}
P.~D. Hoff, ``Bilinear mixed-effects models for dyadic data,'' \emph{J. Am.
  Stat. Assoc.}, vol. 100, no. 469, pp. 286--295, 2005.

\bibitem[Mark et~al.(2019)Mark, Goldsman, Gurbaxani, Keskinocak, and
  Sokol]{mark2019using}
E.~Mark, D.~Goldsman, B.~Gurbaxani, P.~Keskinocak, and J.~Sokol, ``Using
  machine learning and an ensemble of methods to predict kidney transplant
  survival,'' \emph{PLoS ONE}, vol.~14, no.~1, p. e0209068, 2019.

\bibitem[Luck et~al.(2017)Luck, Sylvain, Cardinal, Lodi, and
  Bengio]{luck2017deep}
M.~Luck, T.~Sylvain, H.~Cardinal, A.~Lodi, and Y.~Bengio, ``Deep learning for
  patient-specific kidney graft survival analysis,'' \emph{arXiv preprint
  arXiv:1705.10245}, 2017.

\bibitem[Harrell et~al.(1982)Harrell, Califf, Pryor, Lee, and
  Rosati]{harrell1982evaluating}
F.~E. Harrell, R.~M. Califf, D.~B. Pryor, K.~L. Lee, and R.~A. Rosati,
  ``Evaluating the yield of medical tests,'' \emph{JAMA}, vol. 247, no.~18, pp.
  2543--2546, 1982.

\bibitem[Byrd et~al.(1995)Byrd, Lu, Nocedal, and Zhu]{byrd1995limited}
R.~H. Byrd, P.~Lu, J.~Nocedal, and C.~Zhu, ``A limited memory algorithm for
  bound constrained optimization,'' \emph{SIAM J. Sci. Comput.}, vol.~16,
  no.~5, pp. 1190--1208, 1995.

\bibitem[Cox and Cox(2008)]{cox2008multidimensional}
M.~A.~A. Cox and T.~F. Cox, ``Multidimensional scaling,'' in \emph{Handbook of
  data visualization}.\hskip 1em plus 0.5em minus 0.4em\relax Springer, 2008,
  pp. 315--347.

\bibitem[Huang et~al.(2022)Huang, Soliman, Paul, and Xu]{huang2022mutually}
Z.~Huang, H.~Soliman, S.~Paul, and K.~S. Xu, ``{A mutually exciting latent
  space Hawkes process model for continuous-time networks},'' in \emph{Proc.
  38th Conf. Uncertain. Artif. Intell.}, 2022, pp. 863--873.

\bibitem[Li et~al.(2009)Li, Zhang, and Sindhwani]{li2009non}
T.~Li, Y.~Zhang, and V.~Sindhwani, ``A non-negative matrix tri-factorization
  approach to sentiment classification with lexical prior knowledge,'' in
  \emph{Proc. Jt. Conf. 47th Annu. Meet. ACL Int. Jt. Conf. Nat. Lang. Process.
  AFNLP}, 2009, pp. 244--252.

\end{thebibliography}

\end{document}